\documentclass{ifacconf}

\usepackage{graphicx}      
\usepackage{natbib}        
\usepackage{subfigure}
\usepackage{amsmath}
\usepackage{amssymb}
\usepackage{float}
\usepackage{paralist}
\usepackage{comment}

\begin{document}
\begin{frontmatter}

\title{A Failure Identification and Recovery Framework for a Planar Reconfigurable Cable Driven Parallel Robot\thanksref{footnoteinfo}} 

\thanks[footnoteinfo]{This work is supported in part by the U.S. National Science Foundation under NRI grant 1924721.}

\author[First]{Adhiti Raman} 
\author[First]{Ian Walker} 
\author[First]{Venkat Krovi} 
\author[First]{Matthias Schmid}

\address[First]{Clemson University, Clemson, SC 29607, USA (e-mail: adhitir,iwalker,vkrovi,schmidm@clemson.edu).}

\begin{abstract}                
In cable driven parallel robots (CDPRs), a single cable malfunction usually induces complete failure of the entire robot. However, the lost static workspace (due to failure) can often be recovered through reconfiguration of the cable attachment points on the frame. This capability is introduced by adding kinematic redundancies to the robot in the form of moving linear sliders that are manipulated in a real-time redundancy resolution controller. The presented work combines this controller with an online failure detection framework to develop a complete fault tolerant control scheme for automatic task recovery. This solution provides robustness by combining pose estimation of the end-effector with the failure detection through the application of an Interactive Multiple Model (IMM) algorithm relying only on end-effector information. The failure and pose estimation scheme is then tied into the redundancy resolution approach to produce a seamless automatic task (trajectory) recovery approach for cable failures.  
\end{abstract}


\end{frontmatter}

\section{Introduction}

Cable driven parallel robots (CDPRs) are lightweight mechanisms in which cables replace traditional rigid link actuators. They can be designed and structured to provide manipulability over large workspaces, with a high ratio of stiffness to mass, thus proving useful in application domains such as painting (\cite{gagliardini2015reconfigurable}), inspection (\cite{izard2013integration}), warehousing, and manufacturing, including small-scale and large-scale 3D printing (\cite{chesser2022kinematics, jamshidifar2015adaptive, izard2017large}). Traditional CDPRs have fixed workspace domains and generally invariant maps of workspace and performance quality within those domains. Here, the addition of modularity to the design in the form of geometric reconfigurability offers the advantageous flexibility to improve performance at the end-effector or to move the robot into previously inaccessible workspace domains. This has been demonstrated in numerous recent approaches (\cite{rasheed2019collaborative,zhou2014stiffness,raman2020stiffness,seriani2016modular}). In this study, we demonstrate a new advantage of incorporating modularity by utilizing the offered kinematic redundancies for failure tolerant control.

\begin{figure}[b]
\subfigure[]{\includegraphics[width=0.2\textwidth]{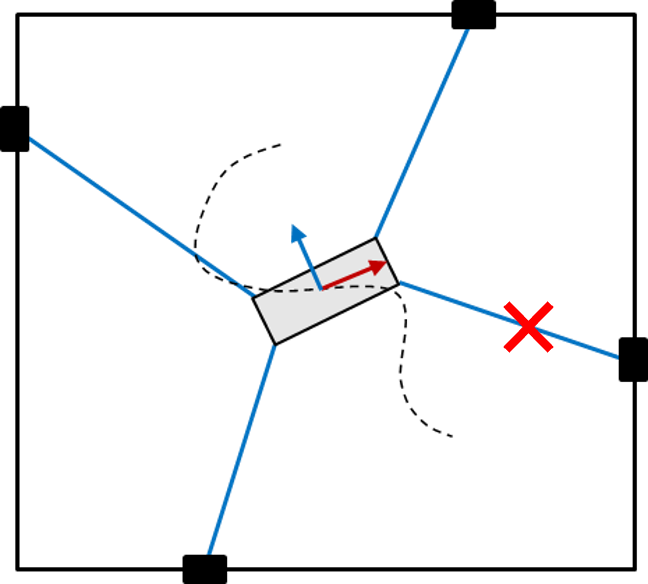}}
\hspace{0.1 cm}
\subfigure[]{\includegraphics[width=0.2\textwidth]{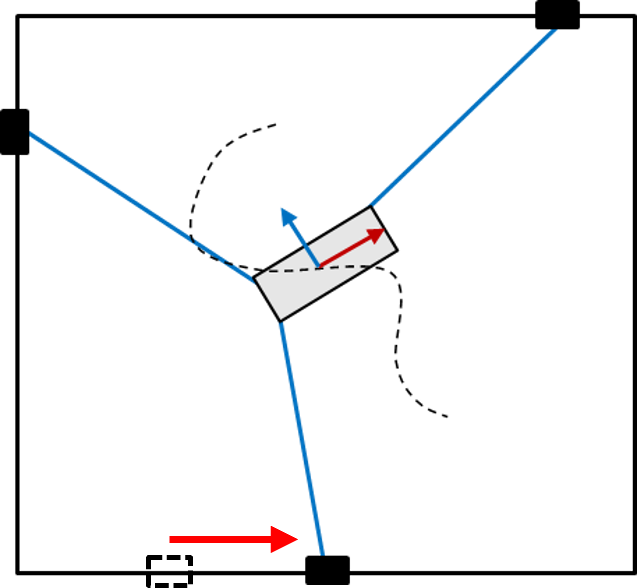}}
\centering
\caption{(a) Cable failure (b) Reconfigurability allows the trajectory tracking to continue}
\label{fig:CableFailureandRecovery}
\end{figure}
Maintaining robot operation despite actuator failure is a rich field of study, especially in aerospace and space robotics where mechanical and controller redundancies are design critical. Despite cable failures being a common problem in CDPRs, however, research on failure tolerant control in this domain remains sparse. As cables can only exert positive tension at the end-effector, this introduces a layer of complexity to more traditional forms of control. The flexibility of the cable introduces unmodeled dynamics and actuator faults which can lead to large uncontrolled tensions in the cables and therefore can cause failures. Here, kinematic redundancy may be employed to overcome cable or actuator failures by utilizing geometric reconfiguration to actively rearrange the attachment points in the structure. This ensures that the static traversable workspace lost through cable failure is quickly recovered. Furthermore, an online-redundancy resolution algorithm ensures that the static workspace travels with the end-effector. The basic idea is depicted in Fig. \ref{fig:CableFailureandRecovery}. 

This study is part of a larger effort that aims at advancing the development of a cooperative robot (cobot) for assisting concrete delivery in tasks such as 3D concrete printing or human directed concrete applications. This cobot is expected to consist of two subsystems: (i) a cable driven parallel manipulator that controls a payload over a large workspace (called the \textit{macro}-manipulator); and (ii) a continuum robot enclosing a concrete delivery tube that provides precisely directed control through congested spaces (called the \textit{micro}-manipulator) (\cite{srivastava20223Dprinting}). The presented work contributes to the development of the first subsystem, for which we have chosen to utilize a kinematically redundant CDPR. We presented the modeling and development of kinematically redundant CDPRs in our previous work (\cite{raman2021wrench}). 

In this paper, we employ geometric redundancies with redundancy resolution control to traverse a trajectory in spite of cable failures. The proposed framework consists of two parts: (i) a failure detection and identification (FDI) section that introduces the use of an Interactive Multiple Model Adaptive Filter; and (ii) a fault tolerant control (FTC) section that utilizes a switching feedforward kinematic controller in tandem with the estimator to form a robust and automatic task recovery scheme.
\section{Related Work}
The use of kinematic redundancy to allow serial rigid-link robots to compensate for joint failures has had a long history~(\cite{visinsky_robot_fault_tolerance, maciejewski_robot_ft}). Early studies accounting for cable failures in CDPRs (\cite{roberts1998inverse}) showed that a static equilibrium could be maintained (upon failure) if the end-effector has been in specific singular configurations. \cite{bosscher2004wrench} noted that there are two kinds of failure modes: cable breakage due to excessive positive cable tensions or slackness due to the lack of any tension. \cite{notash2012failure} looked into additional cable failures, such as stuck actuators (applying a passive restraint on the end-effector) or situations in which the actuator moves but the output is biased. However, these works are limited to kinematic models and CDPRs with ideal and inelastic cables. Other approaches (\cite{passarini2019dynamic} and  \cite{boumann2022simulation}) presented interesting emergency stop strategies (upon cable failure) in cable suspended robots with the focus being on minimizing the safety risk instead of the continued functioning of the robot. 

Although FDI, FTC, or task recovery is a well explored domain in aerospace applications, literature on the same topics in the field of cable driven manipulators is almost non-existent. A fairly popular general technique employed for FDI in other domains is the use of Multi Model Adaptive Estimation (MMAE) filters that assume knowledge of all the possible failure modes that may occur. In this study, we use an Interactive MMAE or IMM for short. IMMs have a rich history in trajectory tracking (\cite{mazor1998interacting}) and behaviour prediction (\cite{gill2019behavior}) as well as in fault detection and identification (\cite{tudoroiu2005fault}). Many of these studies focus on FDI only and do not attempt to integrate controllers for task recovery. One of the few exceptions is the work of \cite{hill2021explicit} where a nonlinear Model Predictive Controller is applied to recover from reaction wheel failures that are identified through an UKF based IMM for satellite maneuvering. Here, we employ a similar framework for FTC.

For our approach, the cables are modeled as elastic springs. Furthermore, we assume that the employed dynamic models are erroneous and do not correspond perfectly to physical reality (e.g., through parametric uncertainties and process noise). We also assume that encoder readings do not correspond accurately to the true position of the end-effector and constitute an unreliable source of information; a reasonable assumption to make when accounting for elasticity in cables. There are several studies addressing the ineffectiveness of relying solely on forward kinematics from cable length measurements to determine end-effector pose, such as works by \cite{le2021cable, korayem2018precise, caverly2016state} where techniques based on Extended (EKF) and Unscented (UKF) Kalman Filtering based on data from payload mounted IMUs have been implemented to mitigate this issue. 

The IMM applies a parallel bank of filters, each corresponding to one of the various failure modes. The output provides an estimate of model corrected pose together with an understanding of the current working mode. Then, the controller computes a mixed joint input to the plant/environment based on real-time information from the IMM. This approach simultaneously detects the fault, provides a corrected estimate, and updates the input, such that the system can robustly cope with sudden failures. This paper serves as a proof-of-concept with key contributions as follows: (i) an Interactive Multi Model Filter is derived and demonstrated as a robust estimator for cable failure detection and identification (FDI) as well as for real-time pose estimation in CDPR applications; and (ii) a first iteration of a proportional-derivative based fault-tolerant control algorithm for task recovery through redundancy resolution post FDI is demonstrated. 

\section{Failure Modes}

\begin{figure}[tb!]
\includegraphics[width=0.45\textwidth]{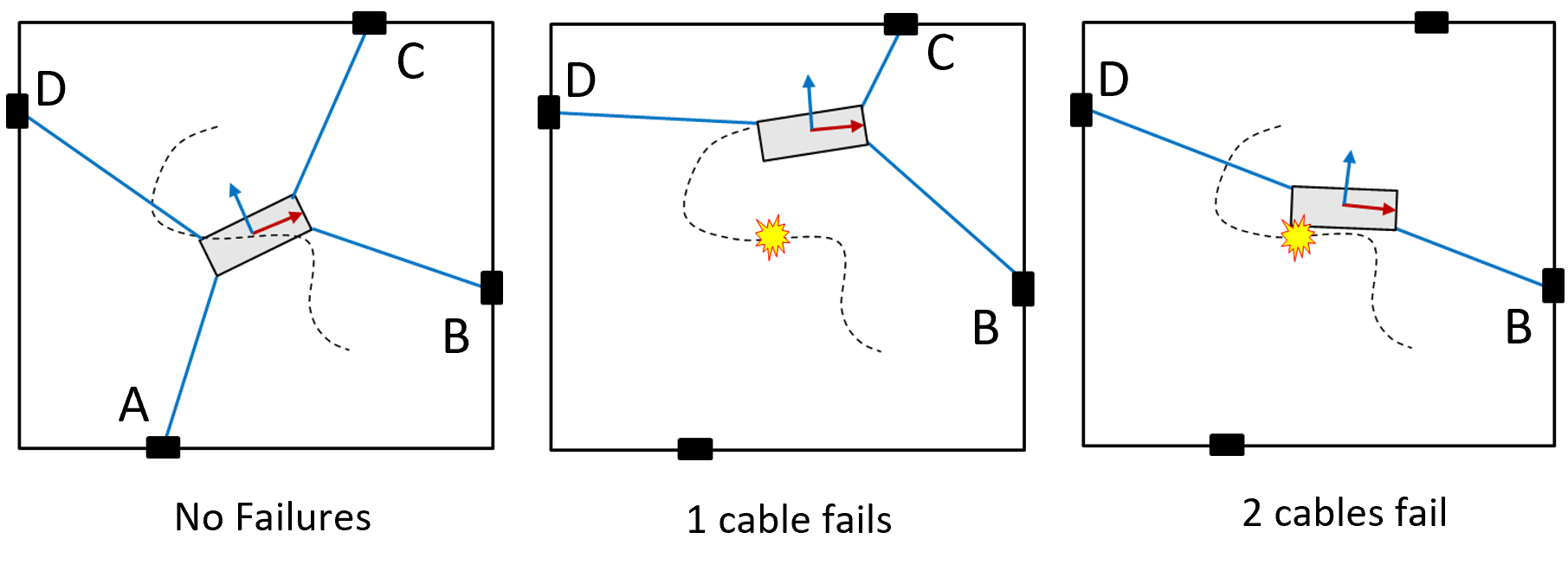}
\centering
\caption{The 3 kinds of failures in a 4 cable kinematically redundant CDPR}
\label{fig:FailureModes}
\end{figure}

The loss of cables in a CDPR reduces the degrees of freedom (DOFs) or the quality of the DOF available at the end-effector. We consider 2 DOFs at the end-effector for the four-cable planar CDPR model in this paper. The total working modes considered can be described by 3 motion models: (i) an over-constrained CDPR with 4 cables; (ii) a fully constrained CDPR with 3 cables; and (iii) an under-constrained CDPR with two cables (Fig~\ref{fig:FailureModes}). 

Therefore there is a total of 7 working modes: no cables have failed (mode 1), only cable A, B, C, or D fail (modes 2-5), cables A and B have failed (mode 6), or cables C and D have failed (mode 7). If any other failure combination occurs, the robot becomes non-navigable for significant areas of the workspace.
\section{CDPR Dynamics}
This section discusses the dynamic model of a planar kinematically redundant CDPR with a 2-DOF platform that is driven by 4 cables and 4 linear actuators housing the cable winch mechanism. The full kinematic model has been presented in our previous work (\cite{raman2021wrench}).
\begin{figure}[tb!]
\includegraphics[width=0.45\textwidth]{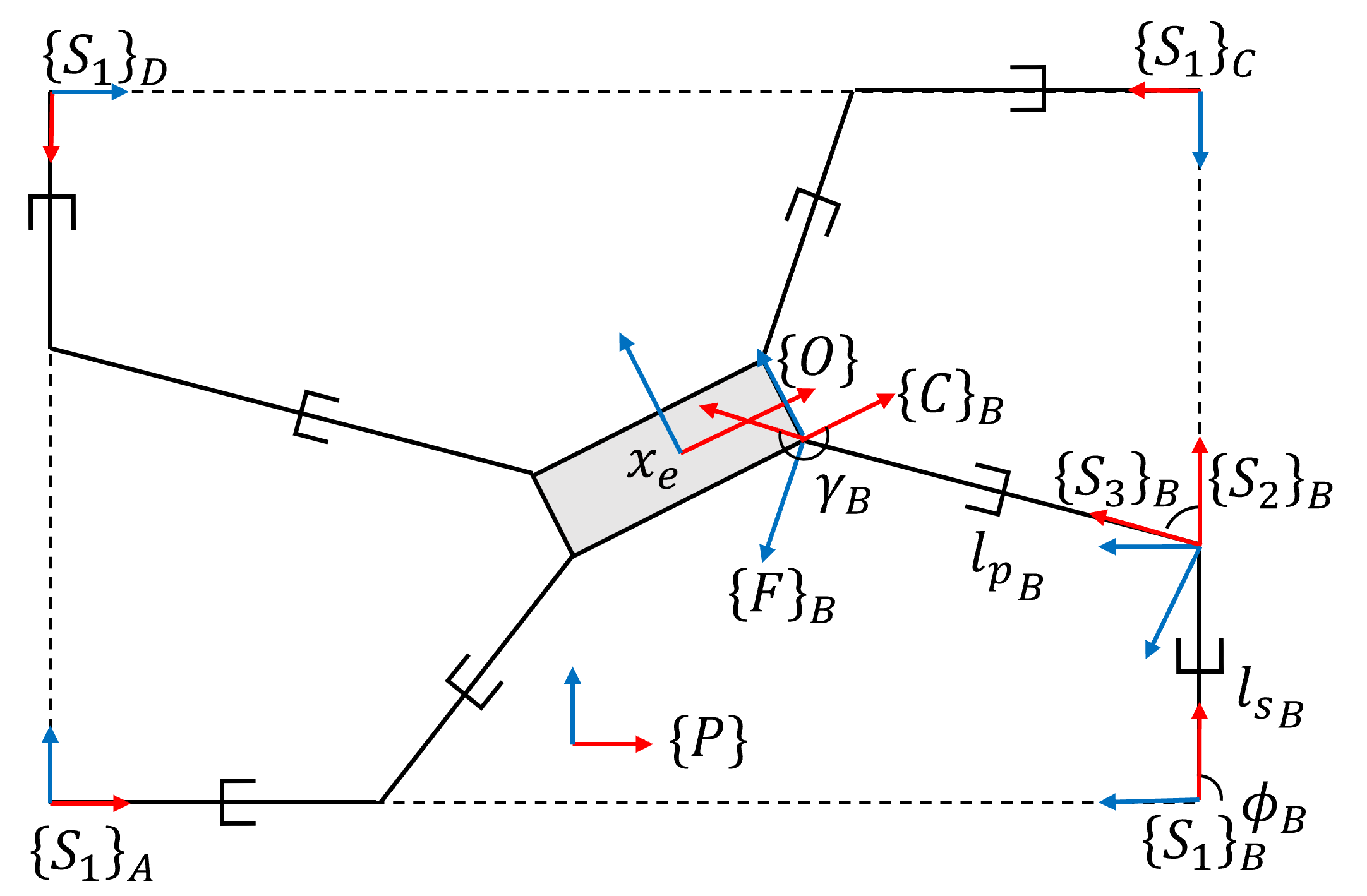}
\centering
\caption{The kinematically redundant 4 cable CDPR}
\label{fig:4PRPR}
\end{figure}
Typical CDPRs require a greater number of cables than the degrees of freedom at the end-effector to be fully constrained. In our model, the cable elements are considered as linear springs with the specific stiffness (stiffness per unit free cable length) of the cable given by $k_{0,i}$. Each cable link forms a prismatic joint between the end-effector $\{F\}_i$ and the base $\{S_3\}_i$ attachment points. The joint length is then given by $l_{p,i}$ which is actuated by a motor-encoder whose position is given by $\theta_i$. This term should ideally be $r\theta_i$ where $r$ is the motor winch radius, but since $r$ is a constant, we subsume the entire term under a generic notation $\theta_i$. Due to the elasticity, $l_{p,i} \not\propto \theta_i$. The cable stiffness at the current time step is given by $k_i=k_{0,i}/l_{p,i}$, where $l_{p,i}$ is determined from inverse kinematics under the assumption of an available end-effector pose (for the controller this comes from forward kinematics). The tension in the cables is given by $k_{i} (l_{p,i} - \theta_i)$, where $l_{p,i} > \theta_i$ to prevent cable sag. The geometric attachment points, $\{S_2\}_i$, are not constant as in traditional CDPRs, but instead move along linear actuators along the $x$-axis of the base frames $\{S_1\}_i$. The longitudinal positions of these attachment points with respect to the base frame are given by $l_{s,i}$. For our model, the linear sliders are considered to be perfect without providing any additional dynamic effects (subject only to limits on velocity). The complete joint input vector can then be expressed as $\textbf{q}=\begin{bmatrix} \textbf{l}_s & \boldsymbol{\theta} \end{bmatrix}$. Note that the lower level control signals translating joint inputs to actuator torque occur independent of the CDPR control loop and are therefore not modeled here.

The cable failure is modeled as a sudden drop in the cable stiffness of the failed joint, i.e., $k_i \to 0$ over a small time interval of $\Delta t = 0.1s$. The full dynamic model of the system can then be derived via the Lagrangian approach as
\begin{equation}
    M \ddot{\mathbf{x}}_e + D\dot{\mathbf{x}}_e = P(\mathbf{x}_e,\mathbf{l}_s) \ K_q \ \Big[ \mathbf{l}_p (\mathbf{x}_e,\mathbf{l}_s) - \boldsymbol{\theta} \Big] 
\end{equation}
where $\textbf{x}_e = [x,y,\phi]$ is the end-effector pose. $P(\textbf{x}_e,\textbf{l}_s)$ is the pulling map or wrench Jacobian. The prismatic length $\textbf{l}_p$ comes from inverse kinematics. Both are dependent on the current pose $\textbf{x}_e$ and geometric attachment points $\textbf{l}_s$ while the joint state stiffness matrix $K_q$ is a diagonal matrix with the elements $(k_1,\dots,k_4)$. The mass and damping matrices are $M$ and $D$ respectively. This formulation ignores the effects of gravity (as the system is planar).   

\section{Motion Models}
Three motion models are employed to describe the seven working modes. For each working mode, the effect of the corresponding cable failure is incorporated in the dynamic model. The states of the end-effector are given by $\textbf{x}= \begin{bmatrix} \textbf{x}_e & \dot{\textbf{x}}_e \end{bmatrix}^T$. The resulting non-linear system model is given by
\begin{align}
    \dot{\textbf{x}}(t) &= \textbf{f}(\textbf{x}(t),\textbf{u}(t),t) + \boldsymbol{\omega} \nonumber \\
    \Tilde{\textbf{y}} &= \textbf{h}(\textbf{x}(t),t) + \textbf{v}
\end{align}
where $\boldsymbol{\omega}$ and $\textbf{v}$ are zero mean Gaussian vector processes accounting for model and measurement noise. The input $\textbf{u}(t)$ to the system is given by the joint input position vector $\textbf{q}(t)$. The non-linear vector functions $\textbf{f}$ and $\textbf{h}$ emerge as
\begin{align}
    \textbf{f}(\textbf{x},\textbf{u}) &= \begin{bmatrix}\dot{\textbf{x}}_e \\ M^{-1} P(\textbf{x}_e,\textbf{l}_s) K_q \big[ \textbf{l}_p(\textbf{x}_e,\textbf{l}_s) - \boldsymbol\theta \big] - M^{-1} D \dot{\textbf{x}}_e  \end{bmatrix} \nonumber \\
    \textbf{h}(\textbf{x}) &= \begin{bmatrix} \textbf{x}_e \\ 0 \end{bmatrix}
\end{align}
where the explicit dependence on time has been dropped for ease of notation. The measurement model assumes that noisy position data is directly available, for instance by separately pre-processing and filtering information from an on-board IMU.  

\section{Interactive Multiple Model Estimation}

\begin{figure}[tb!]
\includegraphics[width=0.45\textwidth]{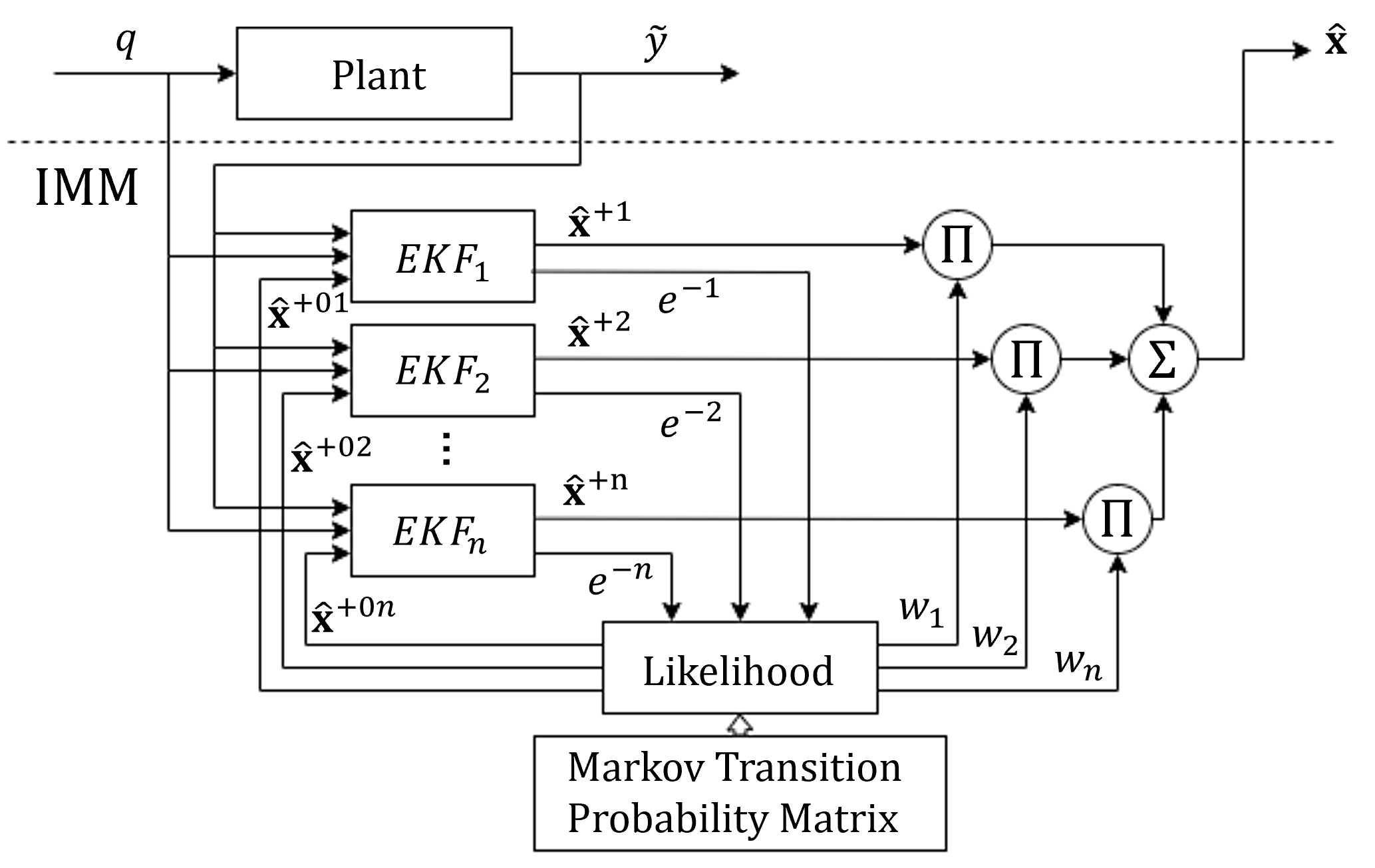}
\centering
\caption{The Interactive Multiple Model (IMM) filter}
\label{fig:IMM}
\end{figure}

An interactive multiple model filter (IMM) is a dynamic estimator (\cite{blom1988interacting}) which can be used when model changes appear suddenly or gradually over time, thus providing means for failure detection. The process flow of an IMM is shown in Fig.~\ref{fig:IMM}. 
In general, the bank of filters in an IMM can be realized by a variety of estimation approaches, i.e., Kalman Filters (KF), Extended Kalman Filters (EKF), Unscented Kalman Filters (UKF), Particle Filters (PF), et cetera. In this initial study, we employ an EKF formulation in which the error dynamics are approximated by a first order Taylor series expansion about the estimated state, $\hat{\textbf{x}}$, i.e.
\begin{equation}
    F = \frac{\partial \textbf{f}(\textbf{x},u)}{\partial \textbf{x}} \Big|_{\mathbf{\hat{x}}}, \quad H = \frac{\partial \textbf{h}(\textbf{x})}{\partial \textbf{x}}\Big|_{\mathbf{\hat{x}}}
\end{equation}
Here, $F$ and $H$ can be determined numerically by a complex step derivative approximation. In this study, the EKF is applied in the discrete time domain; hence, the Euler-Maruyama discretization of the motion model yields 
\begin{align}
    \textbf{x}_k &= \underbrace{\textbf{x}_{k-1} + \textbf{f}(\textbf{x}_{k-1},\textbf{q}_{k-1})\Delta t}_{\textstyle \phi(\textbf{x}_{k-1},\textbf{q}_{k-1})} + \boldsymbol\omega_k \sqrt{\Delta t} \nonumber \\
    \Tilde{\textbf{y}}_k &= \textbf{h}(\textbf{x}_k) + \textbf{v}_k
\end{align}
where $\tilde{\textbf{y}} \in \mathbb{R} ^m$ is the measurement vector, $\boldsymbol\omega_k \sim \mathcal{N}(0,Q_k)$ denotes the process noise with covariance $Q_k \in \mathbb{R}^{n \times n}$, and $\textbf{v}_k \sim \mathcal{N}(0,R_k)$ is the measurement noise with covariance $R_k \in \mathbb{R}^{m \times m}$. The sampling time is denoted by $\Delta t$, and $k$ is the current time step. 

Let $\mathcal{M}$ be the number of all working modes in a CDPR at the $k^{th}$ time step. Then, the input to the filter is the current measurement and mixed state estimate $\hat{\textbf{x}}^{+0}$. The algorithm broadly consists of four stages (\cite{gill2019probabilistic}):
\begin{enumerate}
    \item \textit{Interaction and Mixing:} The weights and the estimates from the last cycle are mixed as per their associated Markov transition probabilities. The predicted probability for the filter to end up in mode $j$ in each cycle, given that it was in the mode $i$ during the previous cycle is given by
\begin{align}
        w^{(i|j)}_k &= \frac{1}{\bar{c}^{j}_k} w^{(i)}_k p_{ij} \nonumber \\
        \bar{c}^{j}_k &= \sum_{i=1}^{\mathcal{M}} w_k^{(i)} p_{ij}
\end{align}
    with $\bar{c}^{j}_k$ being the normalization factor, and where $p_{ij}$ is the Markov transition probability from mode $i$ to mode $j$. This matrix is tuning factor and is determined heuristically. The mixed initial state estimate at the start of the current time step is provided by
    \begin{align}
        \hat{\textbf{x}}^{+0j}_k &= \sum_{i=1}^{\mathcal{M}} w^{(i|j)}_{k} \hat{\textbf{x}}^{+i}_k \nonumber \\
     P^{+0j}_k & = \sum_{i=1}^{\mathcal{M}} w^{(i|j)}_{k} \left[ \left( \hat{\textbf{x}}^{+i}_k - \hat{\textbf{x}}^{+0j}_k\right) \right.  \nonumber \\ & \qquad \qquad \quad \ \ \left. \left( \hat{\textbf{x}}^{+i}_k - \hat{\textbf{x}}^{+0j}_k\right)^T + P^{+i}_k \right]
    \end{align}
    \item \textit{Model Specific Filtering:} The mixed initial estimates are fed into the EKF and processed in two steps, i.e.
    \begin{compactitem}
    \item Propagation:
    \begin{align}
        \hat{\textbf{x}}^{-j}_{k} &= \hat{\textbf{x}}^{+0j}_{k} + \textbf{f}(\hat{\textbf{x}}^{+0j}_{k},\textbf{q}_{k}) \Delta t \nonumber \\
        P^{-j}_{k} &= \Phi_k^j P^{+0j}_{k}\Phi_k^{jT} + Q_k \nonumber \\
        \Phi^j &= I + F_k^j\Delta t
    \end{align}
    \item And update:
    \begin{align}
        K^{(j)}_k &= P^{-j}_{k} H^{jT}_k[H^{j}_k P^{-j}_{k} H^{jT}_k + R_k]^{-1} \nonumber \\
        \hat{\textbf{x}}^{+j}_k &= \hat{\textbf{x}}^{-j}_{k} + K_k[\Tilde{\textbf{y}}_{k} - \textbf{h}(\hat{\textbf{x}}^{-j}_{k})] \nonumber \\
        P^{+j}_{k} &= [I - K_k H_{k}]P^{-j}_k
        \end{align}
    \end{compactitem}
    Here, `$+$' denotes the \textit{a posteriori} estimate whereas `$-$' denotes the \textit{a priori} quantity before the update. The likelihood of a measurement is then given by
    \begin{align}
    p(\Tilde{\textbf{y}}_k| \hat{\textbf{x}}^{-j}_{k}) &= \frac{1}{[\det{(2\pi E^{-j}_k)}]^{1/2}} \nonumber \\
    & \qquad \quad \cdot \exp\Big[{-\frac{1}{2} \textbf{e}^{-jT}_k (E_k^{-j})^{-1} \textbf{e}_k^{-j}} \Big] \nonumber \\
    \text{with} \quad  E^{-j}_k&=H^{j}_k P^{-j}_{k} H^{jT}_k + R_k \nonumber \\
    \textbf{e}^{-j}_k &= \Tilde{\textbf{y}}_{k} - \hat{\textbf{y}}^{-j}_k
    \end{align}
    and where $\textbf{e}_k^{-j}$ is the estimation error. 
    \item \textit{Model Probability Update:} Now, the model probabilities are updated via the likelihood with subsequent normalization, i.e.
    \begin{align}
        w^j_{k} &= w^j_{k-1} p(\Tilde{\textbf{y}}_k| \hat{\textbf{x}}^{-j}_{k}) \nonumber \\
        w^j_{k} &\gets \frac{w^j_{k}}{\sum_{i=1}^{\mathcal{M}} w^j_{k}} 
    \end{align}
    
    \item{\textit{Combination:}} Finally, the updated estimate is given by 
    \begin{align}
    \hat{\textbf{x}}^{+}_k &= \sum_{j=1}^{\mathcal{M}} w^{j}_{k} \hat{\textbf{x}}^{+j}_k \nonumber \\
     P^{+}_k &= \sum_{j=1}^{\mathcal{M}} w^{j}_{k} \left[ \left( \hat{\textbf{x}}^{+j}_k - \hat{\textbf{x}}^{+}_k\right) \left( \hat{\textbf{x}}^{+j}_k - \hat{\textbf{x}}^{+}_k\right)^T + P^{+j}_k \right]
    \end{align}
\end{enumerate}

The final estimate of $\hat{\textbf{x}}$ is not integrated into the overall controller but the weight vector, $w$, plays an important role. The weight vector $w^j_{k}$ displays the importance (probability) associated with each model in the bank. If there are no cables failures, for example, the IMM will determine the first model (mode 1) to have the largest weight. This vector also informs the balance of joint inputs to the controller as described in the next section.

\section{Task Recovery}

We wish to describe a proof-of-concept controller for a CDPR performing 3D printing. Therefore, the primary objective is to maintain end-effector trajectory tracking along a predefined path within the operational workspace of the redundant CDPR despite cable failure. The operational workspace is different from the static workspace. The static workspace is the instantaneous wrench feasible workspace of the CDPR if all the sliders were fixed. The operational workspace is the overall traversable workspace of the CDPR that lies within the total geometric constraints of the system. For the reconfigurable CDPR, the operational workspace lies within the bounds of the blue box in Fig.~\ref{fig:CableFailureandRecovery} (geometric limits of the joints), but we cannot make declarations about the quality and existence of the static workspace (for all redundant configurations) within it with equal ease. However, it is sufficient to state that if the next pose in the trajectory is manipulable and wrench feasible, then it is realizable through the controller.

The desired trajectory of the end-effector is known. The corresponding joint velocities are determined through redundancy resolution. The primary motivation of the task recovery controller is to reduce the tracking error as a primary constraint while maintaining the manipulability and wrench feasibility (i.e., avoiding infeasibility and singularity) of the joint space solutions. In general, this process can be separated into: (i) a task space controller; (ii) a redundancy resolution scheme to maximize manipulability and wrench feasibility; and (iii) a joint level velocity controller.

\subsection{Trajectory Tracking}
The first iteration for the task recovery task space controller was a straightforward discrete-time Proportional-Derivative (PD) based trajectory tracking algorithm, i.e.
\begin{align}
    \textbf{e}_{p}^k &= (\textbf{x}_{e,r}^k - \textbf{x}_{e,f}^k) \nonumber \\
    \textbf{e}_{d}^k &= (\textbf{e}_{p}^k - \textbf{e}_{p}^{k-1})/\Delta t \nonumber \\
    \textbf{x}_{e}^{k+1} &= \textbf{x}_{e,f}^k + \textbf{g}_p  \textbf{e}_{p}^k + \textbf{g}_d \textbf{e}_{d}^k
\end{align}
where $\textbf{x}_{e,r}^k$ is the trajectory point at current time step and $\textbf{x}_{e,f}^k$ comes from forward kinematics. The forward kinematics are determined from a Levenberg-Marquardt minimization of potential energy in the system given the joint states at the current time step. The next desired position for the end-effector, $\textbf{x}_{e}^{k+1}$, is employed to calculate the required joint states. The control gains $\textbf{g}_p$ and $\textbf{g}_d$ are selected as $\begin{bmatrix} 0.6 & 0.6 & 0 \end{bmatrix}$ and $\begin{bmatrix} 0.1 & 0.1 & 0 \end{bmatrix}$, respectively. We do not control motion in the third DOF. 

\subsection{Redundancy Resolution}

For a given end-effector pose, there is an infinite amount of solutions in the joint space for a kinematically redundant CDPR. The optimal joint slider positions are determined from an objective function that seeks to maximize the manipulability ellipsoid at the end-effector. This aids in avoiding singularities as the sliders travel. When a cable fails (second motion model), the manipulability in the third DOF is ignored. The joint sliders are velocity limited to $v_{max}$ and $v_{min}$ in either direction in order to avoid discontinuous motions i.e.,   
\begin{align}
    \textbf{l}^{\ast}_s &= \max_{\textbf{l}_s} \kappa  \nonumber \\
    \textbf{l}_{s}^{k+1} &= \textbf{l}_{s}^{k} +  \Delta \textbf{l}_s \nonumber\\
    \Delta \textbf{l}_s &= \min(\max(\textbf{l}^{\ast}_s - \textbf{l}_{s}^{k},v_{max}\Delta t),v_{min}\Delta t)
\label{eq:14}
\end{align}
where $\kappa$ is a measure of the sensitivity of the pulling map at the desired pose. This is an equivalent property to the measure of manipulability. This value is given by $\kappa = \frac{\min z}{\max z}$, where $z = \text{null}(P)$. The closer $\kappa$ is to 1, the better conditioned the pulling map is. For the final motion model, the DOF at the end-effector reduces to one and the manipulability ellipse reduces to a line making singularity avoidance a non-issue. For this case, since the cables can control only once DOF, the sliders utilize the same controller for trajectory error minimization to determine actuation that will manipulate the end effector in the remaining DOF. 

The joint angles $\boldsymbol\theta$ are determined by first extracting the desired tensions in the cables, i.e., given the desired end-effector position together with $\textbf{l}_{s}^{k+1}$, we determine the minimum positive tensions required to maintain this pose: 
\begin{equation}
    \boldsymbol\tau^{\ast} = \min_{\boldsymbol\tau} \boldsymbol\tau^{T} \boldsymbol\tau \ni P(\textbf{x}_{e}^{k+1},\textbf{l}_{s}^{k+1})\boldsymbol\tau = 0
\end{equation}
The joint angles are then determined in a straightforward fashion by
\begin{equation}
    \boldsymbol\theta = \textbf{l}_{p}^{k+1} - K^{-1}_q\boldsymbol\tau^{\ast}  
\end{equation}
\subsection{Feedforward Kinematic Control}
The task recovery algorithm has a parallel bank of trajectory tracking controllers, each corresponding to a working mode. The controllers use simple kinematic models for the different redundant CDPR motion modes. These inputs are mixed and normalized with the weight vector arising from the IMM, giving precedence to the input corresponding to the detected failure mode. The bank of task- and joint-space controllers each accept the mixed joint state input of the previous time-step and propagate it forward to the next input for each failure mode. The mixed joint inputs are then applied to the dynamic simulation model (plant), and the resulting (noisy) measurements of the end-effector state inform again the state estimation and weight vector of the IMM.

\begin{figure}[tb!]
\includegraphics[width=0.45\textwidth]{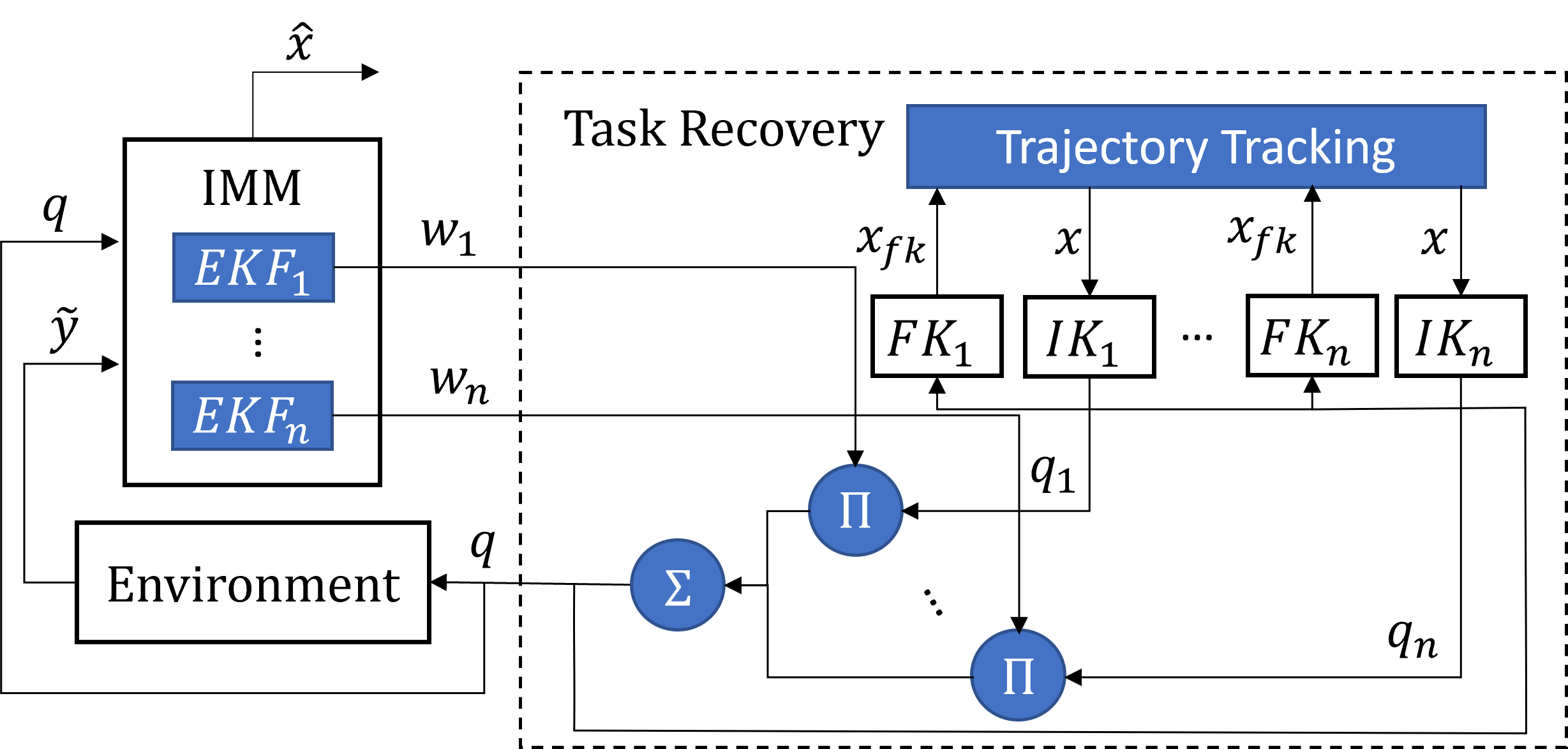}
\centering
\caption{Automatic task recovery through open-loop kinematic control}
\label{fig:taskrecoverycontrol}
\end{figure}
The dynamic simulation of the plant and IMM state estimation utilize a sampling rate different from the controller. The mixing process results in a gradual change of the joint inputs to the plant, thus allowing for recovery if a failure mode is incorrectly detected. The complete algorithmic framework is shown in Fig.~\ref{fig:taskrecoverycontrol}.

\section{Implementation and Results}
\subsection{Simulation Framework}
The plant is a dynamic model created in MATLAB which is perturbed with process noise to mimic unmodeled dynamics. The measurements are simulated though a measurement noise perturbation as well. The cables are considered as linear springs (where stiffness is a function of free cable length) with no mass and no sag while all other model parameters are constants. Cable failure is modeled as a drop in cable stiffness to zero. If the stiffness is zero, the cable will have no tension and thus have no impact on the platform dynamics. The plant and estimation run at 100 Hz while the kinematic controller runs at 10 Hz. 

\subsection{Task Recovery}

\begin{figure}[tb!]
\includegraphics[width=0.45\textwidth]{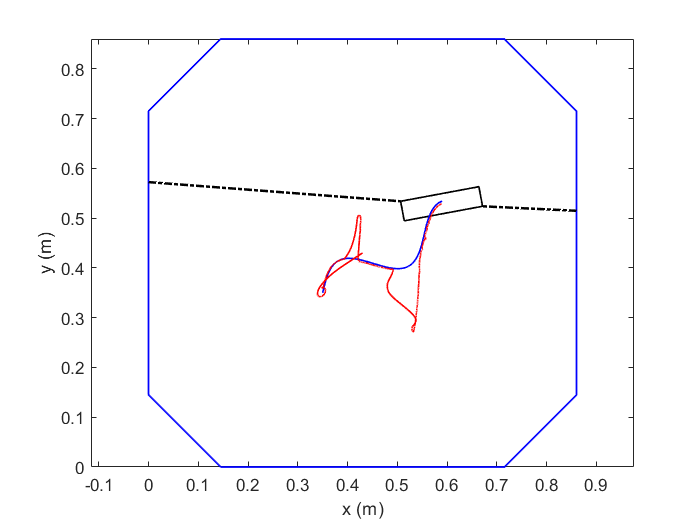}
\centering
\caption{Trajectory tracking and recovery}
\label{fig:taskrecovery}
\end{figure}

\begin{figure}[tb!]
\includegraphics[trim={1cm 0 1.5cm 0.3cm},clip,width=0.45\textwidth]{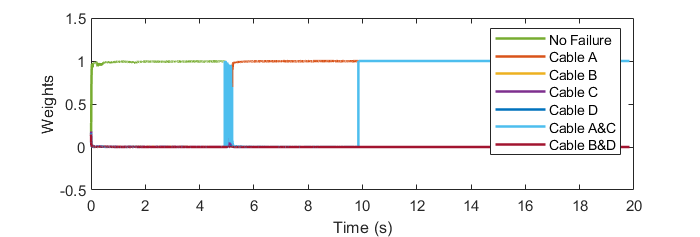}
\centering
\caption{Cable failure identified at 5 and 10 seconds}
\label{fig:weights}
\end{figure}

\begin{figure}[tb!]
\includegraphics[trim={1cm 1cm 1cm 0cm},clip, width=0.45\textwidth]{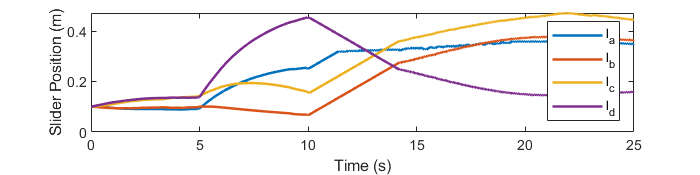}
\centering
\caption{Slider positions during geometric reconfiguration}
\label{fig:geometryreconfig}
\end{figure}

\begin{figure}[tb!]
\includegraphics[trim={1cm 0cm 1.5cm 0cm},clip,width=0.45\textwidth]{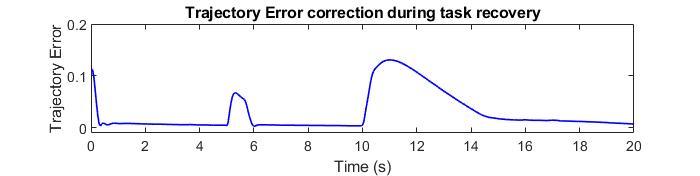}
\centering
\caption{Trajectory error correction during task recovery }
\label{fig:error2}
\end{figure}

Figure~\ref{fig:taskrecovery} illustrates the application of the automatic task recovery algorithm for a case when a first cable breaks at 5 seconds and an additional cable fails at 10 seconds. The red trajectory shows the true path of the end-effector while the blue trajectory depicts the expected path. The red line exhibits a clear break away from the trajectory and subsequent recovery with three cables as the systems returns to and continues from the point of first failure. The second cable failure and subsequent recovery then demonstrates similar behavior further down the trajectory. 
When the cables break, the IMM responds instantly, identifying the correct model and mixing the inputs accordingly, thus leading to recovery. In Fig.~\ref{fig:weights}, the weights vector correctly assigns the largest values to the current working mode. At the five second mark, we can see that the weight vector has correctly identified the cable A failure, followed by the correct identification of the cable C failure at the ten second mark. 
The slider positions change in time to reflect the redundancy resolution during task recovery in Fig.~\ref{fig:geometryreconfig} while the robustness of the task recovery can be seen in Fig.~\ref{fig:error2} as the real and desired trajectories converge.
Working in simulation provides us with access to the ground truth and thus, the estimation error ($\textbf{e} = \textbf{x}_e - \hat{\textbf{x}}_e$), therefore allowing for insight into the stability of the IMM.  
\begin{figure}[tb!]
\includegraphics[trim={1cm 0.8cm 1.5cm 0},clip,width=0.45\textwidth]{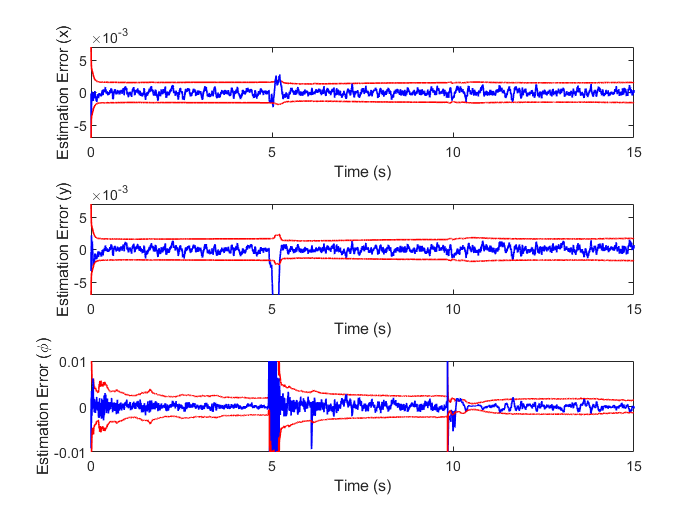}
\centering
\caption{Estimation Error}
\label{fig:error1}
\end{figure}
\subsection{Drawbacks}
When the desired trajectory lies outside of the new static workspace that forms just after cable failure, the robot sometimes struggles to recover the trajectory tracking. This is because the robot may move close to singularity while attempting to minimize the trajectory error too quickly. Sometimes cable failure can also lead to cable slackness. While this property is not explicitly modeled in the system dynamics, the destabilizing effect of this can be identified when the value of spool angles, $\theta$, is too high (at the instant of failure) to converge to a forward kinematics solution that lies within the new static workspace. For 3D printing tasks, we also require the end-effector to return to the point of break. These drawbacks will be addressed in our next work. 

\section{Conclusion}
In this study, we demonstrate for the first time a proof-of-concept approach to failure detection and identification (FDI) with fault tolerant control (FTC) for kinematically redundant CDPRs by automating the cycle of detection and recovery. CDPRs offer a unique advantage by being able to utilize geometric reconfiguration in overcoming the consequences of cable failure. Failure modes are detected/estimated via an Interactive Multiple Model (IMM) algorithm that relies on a bank of parallel Extended Kalman Filters. The actual joint state inputs are accomplished by mixing the simultaneously calculated inputs from a parallel bank of pre-developed trajectory tracking controllers utilizing the IMM weight vector. While IMM approaches have been employed in dynamic fault detection and identification for aerospace applications, its use for fault detection in the robotics domain in general is sparse and truly unique for the cable driven robotics community. Furthermore, the recovery strategy incorporating both redundancy resolution and geometric reconfiguration for task recovery with the utilization of the IMM weights for control mixing is a first in this domain. In addition to providing a novel research idea, this also serves as a vehicle demonstrating the powerful advantage of adding kinematic redundancies to CDPRs. This work is meant to be a first study of a concept that requires advancement and refinement in future work, for instance by incorporating a closed loop dynamic controller together with a motion planning strategy to recover the trajectory immediately from the point of failure. Further exploration is suggested on how to incorporate redundant actuation to minimize dynamic effects of cable failure to avoid cable and object collision. 

\begin{ack}
This work is supported in part by the U.S. National Science Foundation under NRI grant 1924721.
\end{ack}

\bibliography{ifacconf}             

\begin{thebibliography}{27}
\providecommand{\natexlab}[1]{#1}
\providecommand{\url}[1]{\texttt{#1}}
\providecommand{\urlprefix}{URL }
\expandafter\ifx\csname urlstyle\endcsname\relax
  \providecommand{\doi}[1]{doi:\discretionary{}{}{}#1}\else
  \providecommand{\doi}{doi:\discretionary{}{}{}\begingroup
  \urlstyle{rm}\Url}\fi

\bibitem[{Blom and Bar-Shalom(1988)}]{blom1988interacting}
Blom, H.A. and Bar-Shalom, Y. (1988).
\newblock The interacting multiple model algorithm for systems with markovian
  switching coefficients.
\newblock \emph{IEEE transactions on Automatic Control}, 33(8), 780--783.

\bibitem[{Bosscher and Ebert-Uphoff(2004)}]{bosscher2004wrench}
Bosscher, P. and Ebert-Uphoff, I. (2004).
\newblock Wrench-based analysis of cable-driven robots.
\newblock In \emph{IEEE International Conference on Robotics and Automation,
  2004. Proceedings. ICRA'04. 2004}, volume~5, 4950--4955. IEEE.

\bibitem[{Boumann and Bruckmann(2022)}]{boumann2022simulation}
Boumann, R. and Bruckmann, T. (2022).
\newblock Simulation and model-based verification of an emergency strategy for
  cable failure in cable robots.
\newblock In \emph{Actuators}, volume~11, 56. MDPI.

\bibitem[{Caverly and Forbes(2016)}]{caverly2016state}
Caverly, R.J. and Forbes, J.R. (2016).
\newblock State estimator design for a single degree of freedom cable-actuated
  system.
\newblock \emph{Journal of the Franklin Institute}, 353(18), 4845--4869.

\bibitem[{Chesser et~al.(2022)Chesser, Wang, Vaughan, Lind, and
  Post}]{chesser2022kinematics}
Chesser, P.C., Wang, P.L., Vaughan, J.E., Lind, R.F., and Post, B.K. (2022).
\newblock Kinematics of a cable-driven robotic platform for large-scale
  additive manufacturing.
\newblock \emph{Journal of Mechanisms and Robotics}, 14(2).

\bibitem[{Gagliardini et~al.(2015)Gagliardini, Caro, Gouttefarde, Wenger, and
  Girin}]{gagliardini2015reconfigurable}
Gagliardini, L., Caro, S., Gouttefarde, M., Wenger, P., and Girin, A. (2015).
\newblock A reconfigurable cable-driven parallel robot for sandblasting and
  painting of large structures.
\newblock In \emph{Cable-Driven Parallel Robots}, 275--291. Springer.

\bibitem[{Gill et~al.(2019)Gill, Pisu, Krovi, and Schmid}]{gill2019behavior}
Gill, J.S., Pisu, P., Krovi, V.N., and Schmid, M.J. (2019).
\newblock Behavior identification and prediction for a probabilistic risk
  framework.
\newblock In \emph{Dynamic Systems and Control Conference}, volume 59155,
  V002T25A004. American Society of Mechanical Engineers.

\bibitem[{Gill(2019)}]{gill2019probabilistic}
Gill, J.S. (2019).
\newblock \emph{Probabilistic Framework for Behavior Characterization of
  Traffic Participants Enabling Long Term Prediction}.
\newblock Ph.D. thesis, Clemson University.

\bibitem[{Hill et~al.(2021)Hill, Gadsden, and Biglarbegian}]{hill2021explicit}
Hill, E., Gadsden, S.A., and Biglarbegian, M. (2021).
\newblock Explicit nonlinear mpc for fault tolerance using interacting multiple
  models.
\newblock \emph{IEEE Transactions on Aerospace and Electronic Systems}, 57(5),
  2784--2794.

\bibitem[{Izard et~al.(2017)Izard, Dubor, Herv{\'e}, Cabay, Culla, Rodriguez,
  and Barrado}]{izard2017large}
Izard, J.B., Dubor, A., Herv{\'e}, P.E., Cabay, E., Culla, D., Rodriguez, M.,
  and Barrado, M. (2017).
\newblock Large-scale 3d printing with cable-driven parallel robots.
\newblock \emph{Construction Robotics}, 1(1), 69--76.

\bibitem[{Izard et~al.(2013)Izard, Gouttefarde, Baradat, Culla, and
  Sall{\'e}}]{izard2013integration}
Izard, J.B., Gouttefarde, M., Baradat, C., Culla, D., and Sall{\'e}, D. (2013).
\newblock Integration of a parallel cable-driven robot on an existing building
  fa{\c{c}}ade.
\newblock In \emph{Cable-driven parallel robots}, 149--164. Springer.

\bibitem[{Jamshidifar et~al.(2015)Jamshidifar, Fidan, Gungor, and
  Khajepour}]{jamshidifar2015adaptive}
Jamshidifar, H., Fidan, B., Gungor, G., and Khajepour, A. (2015).
\newblock Adaptive vibration control of a flexible cable driven parallel robot.
\newblock \emph{IFAC-PapersOnLine}, 48(3), 1302--1307.

\bibitem[{Korayem et~al.(2018)Korayem, Yousefzadeh, and
  Kian}]{korayem2018precise}
Korayem, M., Yousefzadeh, M., and Kian, S. (2018).
\newblock Precise end-effector pose estimation in spatial cable-driven parallel
  robots with elastic cables using a data fusion method.
\newblock \emph{Measurement}, 130, 177--190.

\bibitem[{Le~Nguyen and Caverly(2021)}]{le2021cable}
Le~Nguyen, V. and Caverly, R.J. (2021).
\newblock Cable-driven parallel robot pose estimation using extended kalman
  filtering with inertial payload measurements.
\newblock \emph{IEEE Robotics and Automation Letters}, 6(2), 3615--3622.

\bibitem[{Maciejewski(1990)}]{maciejewski_robot_ft}
Maciejewski, A.A. (1990).
\newblock Fault tolerant properties of kinematically redundant manipulators.
\newblock In \emph{IEEE International Conference on Robotics and Automation},
  638--642.

\bibitem[{Mazor et~al.(1998)Mazor, Averbuch, Bar-Shalom, and
  Dayan}]{mazor1998interacting}
Mazor, E., Averbuch, A., Bar-Shalom, Y., and Dayan, J. (1998).
\newblock Interacting multiple model methods in target tracking: a survey.
\newblock \emph{IEEE Transactions on aerospace and electronic systems}, 34(1),
  103--123.

\bibitem[{Notash(2012)}]{notash2012failure}
Notash, L. (2012).
\newblock Failure recovery for wrench capability of wire-actuated parallel
  manipulators.
\newblock \emph{Robotica}, 30(6), 941--950.

\bibitem[{Passarini et~al.(2019)Passarini, Zanotto, and
  Boschetti}]{passarini2019dynamic}
Passarini, C., Zanotto, D., and Boschetti, G. (2019).
\newblock Dynamic trajectory planning for failure recovery in cable-suspended
  camera systems.
\newblock \emph{Journal of Mechanisms and Robotics}, 11(2).

\bibitem[{Raman et~al.(2020)Raman, Schmid, and Krovi}]{raman2020stiffness}
Raman, A., Schmid, M., and Krovi, V. (2020).
\newblock Stiffness modulation for a planar mobile cable-driven parallel
  manipulators via structural reconfiguration.
\newblock In \emph{International Design Engineering Technical Conferences and
  Computers and Information in Engineering Conference}, volume 83990,
  V010T10A054. American Society of Mechanical Engineers.

\bibitem[{Raman et~al.(2021)Raman, Schmid, and Krovi}]{raman2021wrench}
Raman, A., Schmid, M., and Krovi, V.N. (2021).
\newblock Wrench analysis of kinematically redundant planar cdprs.
\newblock In \emph{International Conference on Cable-Driven Parallel Robots},
  90--104. Springer.

\bibitem[{Rasheed(2019)}]{rasheed2019collaborative}
Rasheed, T. (2019).
\newblock \emph{Collaborative Mobile Cable-Driven Parallel Robots}.
\newblock Ph.D. thesis, {\'E}cole centrale de Nantes.

\bibitem[{Roberts et~al.(1998)Roberts, Graham, and
  Lippitt}]{roberts1998inverse}
Roberts, R.G., Graham, T., and Lippitt, T. (1998).
\newblock On the inverse kinematics, statics, and fault tolerance of
  cable-suspended robots.
\newblock \emph{Journal of Robotic Systems}, 15(10), 581--597.

\bibitem[{Seriani et~al.(2016)Seriani, Gallina, and
  Wedler}]{seriani2016modular}
Seriani, S., Gallina, P., and Wedler, A. (2016).
\newblock A modular cable robot for inspection and light manipulation on
  celestial bodies.
\newblock \emph{Acta Astronautica}, 123, 145--153.

\bibitem[{Srivastava et~al.(2022)Srivastava, Ammons, Peerzada, Krovi,
  Rangaraju, and Walker}]{srivastava20223Dprinting}
Srivastava, M., Ammons, J., Peerzada, A., Krovi, V., Rangaraju, P., and Walker,
  I.D. (2022).
\newblock 3d printing of concrete with a continuum robot hose using variable
  curvature kinematics.
\newblock In \emph{IEEE International Conference on Robotics and Automation
  (ICRA)}. IEEE.

\bibitem[{Tudoroiu and Khorasani(2005)}]{tudoroiu2005fault}
Tudoroiu, N. and Khorasani, K. (2005).
\newblock Fault detection and diagnosis for satellite's attitude control system
  (acs) using an interactive multiple model (imm) approach.
\newblock In \emph{Proceedings of 2005 IEEE Conference on Control Applications,
  2005. CCA 2005.}, 1287--1292. IEEE.

\bibitem[{Visinsky et~al.(1994)Visinsky, Cavallaro, and
  Walker}]{visinsky_robot_fault_tolerance}
Visinsky, M.L., Cavallaro, J.R., and Walker, I.D. (1994).
\newblock Robotic fault detection and fault tolerance: A survey.
\newblock \emph{Reliability Engineering and System Safety}, 46(2), 139--158.

\bibitem[{Zhou et~al.(2014)Zhou, Jun, and Krovi}]{zhou2014stiffness}
Zhou, X., Jun, S.k., and Krovi, V. (2014).
\newblock Stiffness modulation exploiting configuration redundancy in mobile
  cable robots.
\newblock In \emph{2014 IEEE International Conference on Robotics and
  Automation (ICRA)}, 5934--5939. IEEE.

\end{thebibliography}
                                                   







\end{document}